\documentclass[runningheads]{llncs}
\usepackage{graphicx}
\usepackage{amsmath,amssymb} 
\usepackage{color}
\usepackage[symbol]{footmisc}
\usepackage{booktabs}
\usepackage{multirow}
\usepackage{threeparttable}

\usepackage{cvpr_eso}
\usepackage[width=122mm,left=12mm,paperwidth=146mm,height=193mm,top=12mm,paperheight=217mm]{geometry}

\begin{document}

\title{Weakly-Supervised Multi-Level Attentional Reconstruction Network for Grounding Textual Queries in Videos} 


\titlerunning{Weakly-Supervised Multi-Level Attentional Reconstruction Network}
%
\author{Yijun Song\inst{1}\thanks{Work done while Yijun Song was a Research Intern with Tencent AI Lab.}\and
Jingwen Wang\inst{2}$^{\dag}$  \and Lin Ma\inst{2}$^{\dag}$
\and Zhou Yu\inst{1} \and Jun Yu\inst{1}}
\authorrunning{Y. Song, J. Wang, L. Ma, Z. Yu and J. Yu.}
%
\institute{Hangzhou Dianzi University, Hangzhou, China \and
Tencent AI Lab, Shenzhen, China \\
\email{\{yijunsong.0377, jaywongjaywong, forest.linma\}@gmail.com}\\
\email{\{yuz, junyu\}@hdu.edu.cn}}

\maketitle
\renewcommand{\thefootnote}{\dag}
\footnotetext{Corresponding authors.}

\begin{abstract}
The task of temporally grounding textual queries in videos is to localize one video segment that semantically corresponds to the given query. Most of the existing approaches rely on segment-sentence pairs (temporal annotations) for training, which are usually unavailable in real-world scenarios. In this work we present an effective weakly-supervised model, named as Multi-Level Attentional Reconstruction Network (MARN), which only relies on video-sentence pairs during the training stage. The proposed method leverages the idea of attentional reconstruction and directly scores the candidate segments with the learnt proposal-level attentions. Moreover, another branch learning clip-level attention is exploited to refine the proposals at both the training and testing stage. We develop a novel proposal sampling mechanism to leverage intra-proposal information for learning better proposal representation and adopt 2D convolution to exploit inter-proposal clues for learning reliable attention map. Experiments on Charades-STA and ActivityNet-Captions datasets demonstrate the superiority of our MARN over the existing weakly-supervised methods.


\end{abstract}

\section{Introduction}
\label{sec:introduction}
\begin{figure}[t]
\begin{center}
\includegraphics [width=0.7\linewidth]{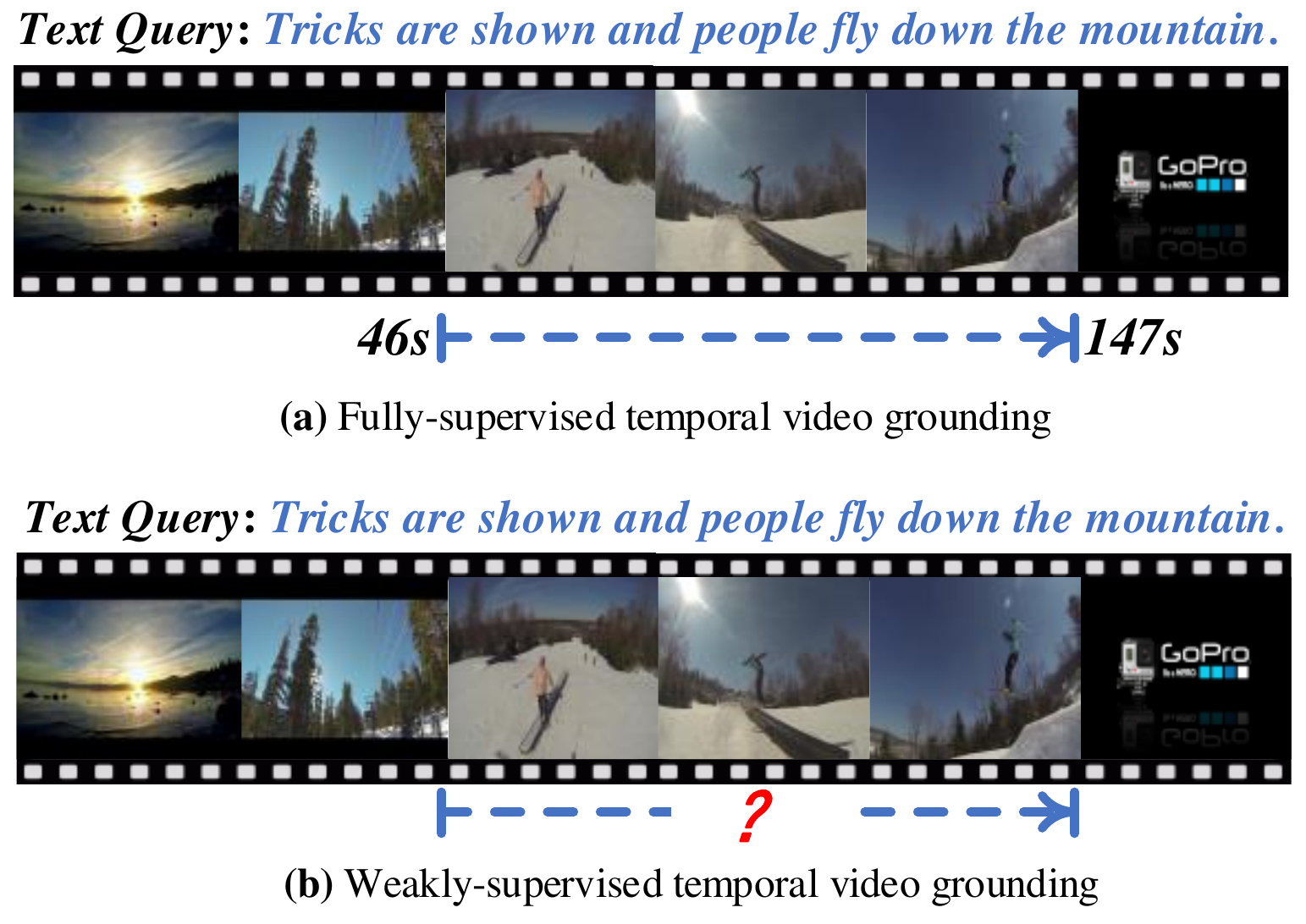}
\caption{The task of fully-supervised temporal video grounding (a) and the task of weakly-supervised temporal video grounding (b). For the weakly-supervised setting, there is no associated temporal annotation (start/end timestamps) with the query language during the training stage.}
\label{fig:introduction}
\end{center}
\end{figure}

Temporal action localization \cite{liu2016ssd,zhao2017temporal,buch2017sst,lin2018bsn,lin2019bmn} has long been explored, where actions of interest are temporally localized in untrimmed videos. The following works extend this task and facilitate localizing the target video segment via a given textual query \cite{anne2017localizing,gao2017tall,chen2018tgn,yuan2019scd} (Figure\ref{fig:introduction} (a)). Recently, the task is further extended to weakly-supervised scenario, where only video-sentence annotation is available at the training stage \cite{mithun2019tga,gao2019wslln,tan2019wman,lin2019scn,chen2020look}. The task, namely the weakly-supervised temporal grounding of textual queries in videos (Figure \ref{fig:introduction} (b)), does not require heavy human annotation and thus enjoys good scalability to real-world datasets, such as the large-scale online videos with captions uploaded by users. Such weakly-supervised task is challenging as it requires certain models to automatically filter out irrelevant content while localizing the matched video segment with only video-level guidance.

For the weakly-supervised temporal video grounding task, some approaches rely on learning frame-level attention to obtain global video representation, followed by learning matching relationship between videos and texts \cite{mithun2019tga,tan2019wman}. However, it is non-trivial and indirect to obtain good proposal-level predictions from frame-level predictions (attentions) at the inference stage. Therefore, in \cite{mithun2019tga}, sliding windows with scales of 128 and 256 frames are adopted to summarize the frame-level attentions and make proposal predictions. In \cite{tan2019wman}, a tricky LogSumExp pooling operation is adopted to determine the relevance between each proposal with the query. Both methods could bring misalignment between training and testing. Some other methods train a proposal selection module by sampling high-score proposals during each training iteration and calculating proposal-sentence matching loss \cite{gao2019wslln,chen2020look,lin2019scn}. We argue that the above mentioned approaches are indirect and insufficient to correlate the proposal scoring process with the video-level supervision.

To enable deeper coupling between the proposal scoring procedure and the video-level supervision, we introduce a novel mechanism called \textbf{Attentional Reconstruction} for weakly-supervised grounding of textual queries in videos. As shown in Figure \ref{fig:method}, Attentional Reconstruction explicitly calculates proposal-level attentions corresponding to a given textual query, then yields a global video representation by attentively summarizing all the proposals. As we want to highlight the important proposals corresponding to the given query, a query reconstruction is performed based on the obtained global video representation. The underlying hypothesis is that the semantically-related proposal with the query should be highly attended when the attended global video representation is able to well reconstruct the given query. With such a mechanism, the proposal scoring process and the final video-level supervision can be deeply coupled. Also, the inference process is greatly simplified. At the inference stage, the proposals are directly ranked by the learnt attention scores.

As proposals in a video exhibit extremely high variance regarding proposal scale (duration), it is better to perform proper modeling and maintain discriminative and learnable features. The commonly-adopted ways are simple mean pooling or max pooling to aggregate features in order to obtain fix-length proposal representation. Inspired by BMN \cite{lin2019bmn}, we propose to use an efficient proposal representation module to overcome the modeling difficulty imposed by the variant proposal scales. Specifically, we first map the original video features to a lower dimensionality, and then construct a proposal sampling weight matrix, which can greatly speed up the feature interpolation process. Then, 3D convolutions are exploited to summarize the intra-proposal features and obtain fix-length proposal representation. We empirically demonstrate that our proposal representation shows much better grounding performance than the simple mean/max-pooling. Afterwards, we use stacked 2D convolutions which captures inter-proposal relationships for calculating proposal-level attentions.

The generated proposal map is further integrated with the given textual query, and is used to calculate the query-driven proposal-level attention map. Following similar attentional reconstruction idea, another branch focusing on frame-level information is constructed to obtain the corresponding frame-level attention map, which is further leveraged to refine the proposal map at both the training stage and the inference stage.

Our contributions can be summarized as follows.
\begin{itemize}
    \item We introduce a novel attentional reconstruction mechanism to the task of weakly-supervised grounding textual queries in videos. It greatly simplifies the proposal scoring process at both the training stage and the inference stage. It also aligns the training and inference process. We further extend the idea to leverage multi-level information (e.g., proposal-level and clip-level clues). The multi-level framework can benefit both the training and inference stages.
    \item We introduce a learnable proposal representation module that is capable of tacking the different proposal scales and generating more discriminative proposal features with both intra-proposal and inter-proposal interaction.
    \item The conducted extensive experiments demonstrate that the proposed Multi-Level Attentional Reconstruction Network (MARN) surpasses the existing state-of-the-art weakly-supervised methods with a clear margin.
\end{itemize}

\section{Related Work}

\subsection{Video Grounding}
The task of temporal video grounding is to localize the matched video segment corresponding to a given textual query. It is a challenge task since it not only requires to find sentence-specific video segment, but also to get precise proposal boundary. To solve this task, \cite{gao2017tall} proposed a framework which takes sliding-window proposals as input, which are further fused with the given query to obtain multi-model representation. Alignment scores are predicted for proposals and location offsets are produced to refine the temporal boundary. Based on this framework proposed by \cite{gao2017tall}, \cite{wu2018mcf} proposed a Multi-modal Circulant Fusion (MCF) to integrate multiple interactions between different modalities. \cite{liu2018attentive} designed a memory attention network to dynamically process the context features. Taking it as a retrieval task, \cite{anne2017localizing} divides the whole video into fix-length segments and extracts candidate moments by consecutive segments. The visual and textual modalities are projected into the same space and the squared distance is minimized. Based on SST framework \cite{buch2017sst}, \cite{chen2018tgn} introduced a frame-by-word attentional interaction mechanism which dynamically leverages frame-level features and word-level features. The grounding results can be produced in one single pass. Based on \cite{chen2018tgn}, \cite{wang2019cbp} further proposed an end-to-end boundary-aware module to predict semantic boundaries, which is utilized to modulate the anchor scores. Also, a self-attention -based contextual aggregation mechanism is provided to improve the grounding performance \cite{wang2019cbp}. \cite{xu2019multilevel} generates query-specific proposals from videos for better selection and involves an early fusion -based retrieval strategy. \cite{liu2018temporal} designed a temporal modular network to exploit underlying language structure. In \cite{chen2019semantic}, a visual concept based approach was proposed to generate proposals, followed by proposal evaluation and refinement. \cite{wang2019language,hahn2019tripping} utilizes reinforcement learning to find the corresponding segments to the queries. Semantic activity concepts are exploited in \cite{ge2019mac} to enhance the temporal grounding task. Based on SSD framework \cite{liu2016ssd}, \cite{yuan2019scd} proposed a semantic conditioned dynamic modulation (SCDM) mechanism to correlate and compose the sentence-specific video content over time.

\subsection{Weakly-Supervised Video Grounding}
Weakly-supervised video grounding is to predict the most semantically matched temporal proposal without temporal segment annotation. It is a more challenging task since the segment localization information is unavailable and we have to use video-sentence alignment annotations instead of segment-sentence pairs during the training stage. To tackle this task, \cite{mithun2019tga} proposed a Text-Guided Attention (TGA) method which trains a clip-query attention to match query related video among batch-wise queries and videos, and calculate the scores for pre-defined proposals by summarizing the clip-level attentions. \cite{tan2019wman} proposed a Moment Alignment Network (wMAN) to learn frame-query alignment by co-attention mechanism. LogSumExp (LSE) pooling similarity metric \cite{lee2018logsumexp} is utilized to score proposals at the inference stage. \cite{gao2019wslln} took proposal-level features as input, they measured video-text alignment and used it to generate pseudo GT to train segment-text alignment. \cite{chen2020look} provided a two-stage model. They adopt sliding window fashion to obtain proposals as input, and train the proposal-text alignment in a coarse-to-fine manner. \cite{lin2019scn} proposed a semantic completion module that predicts the important words in a query by sampling top-K proposals. To train the proposal selection module, a proposal reward is calculated based on the language loss. Most of the previous weakly-supervised approaches either rely on frame-level attentions to generate proposal-level predictions \cite{mithun2019tga,tan2019wman} or require proposal sampling \cite{gao2019wslln,lin2019scn} to optimize the model. There could exist misalignment between the training and the inference for those methods. Our proposed MARN model minimizes the gap between the training and the inference.


\section{Weakly-Supervised Multi-level Attentional Reconstruction Network}
\begin{figure}[t]
\centering
\includegraphics[width=1.0\textwidth]{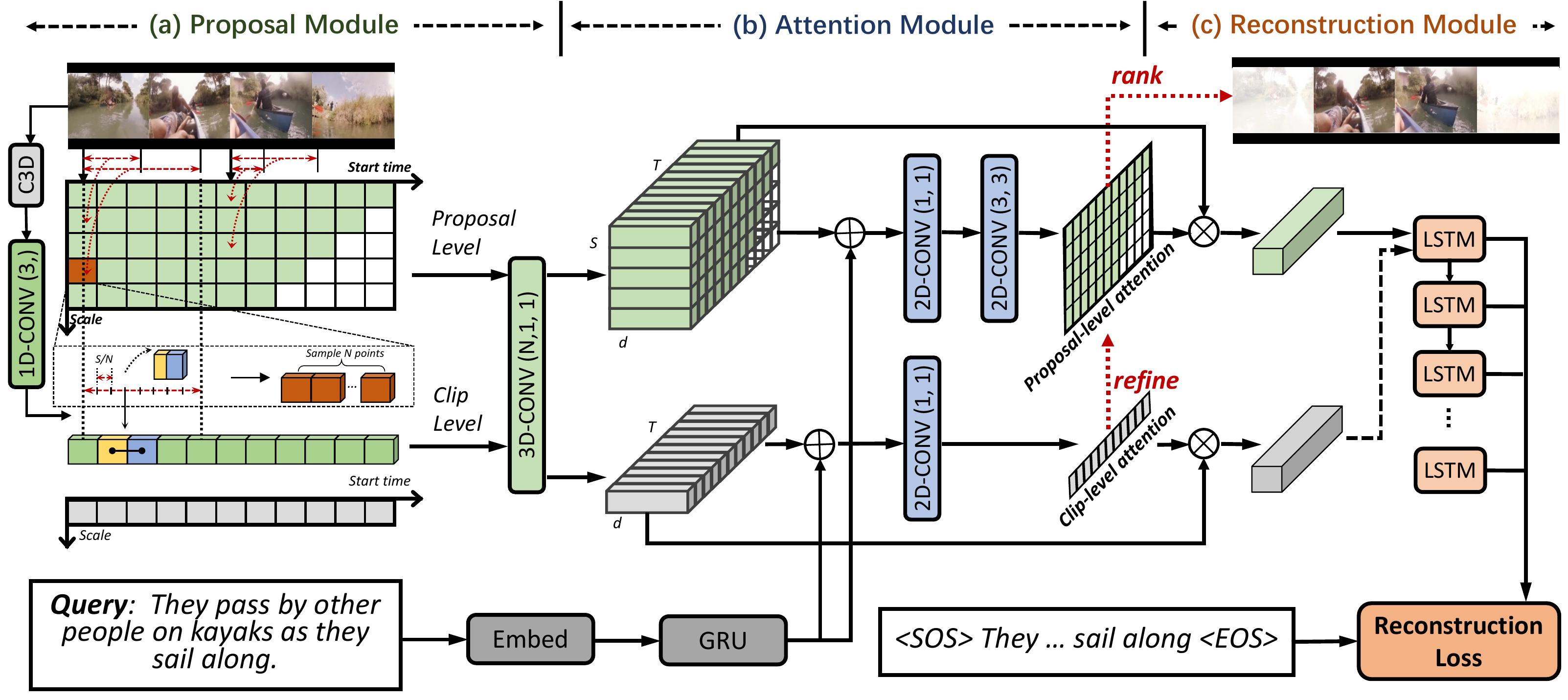}
\caption{The main framework of our proposed method: Multi-Level Attentional Reconstruction Network (MARN). The model consists of three main components: a proposal module (a) to effectively represent proposals of variant scales, an attention module (b) to calculate proposal-level (clip-level) attention scores, and a reconstruction module (c) to leverage query reconstruction to regularize the learnt attentions. The three components are coupled together and can be trained in an end-to-end manner. The candidate proposals can be directly ranked by the learnt attention scores at the inference stage.}
\label{fig:method}
\end{figure}

In this section, we introduce our Multi-Level Attentional Reconstruction Network (MARN) for temporally grounding textual queries in videos, under the weakly-supervised setting. Our model consists of three main components, namely, a proposal module, an attention module, and a reconstruction module, as shown in Fig. \ref{fig:method}. The three components are deeply integrated and enable end-to-end optimization.

\noindent\textbf{Problem Formulation.} Given a query, the video grounding network aims to localize the most relevant segment in the video. The task can be formulated as $\mathbb{P} = VGN(V, Q)$. $V=\{v_k\}_{k=1}^K$ is a video containing $K$ frames. $Q=\{w_m\}_{m=1}^M$ is a query containing $M$ words. $\mathbb{P}=\{(t_s, t_e)_i, s_i\}$ is the obtained ranked proposals by $VGN$, where $t_s, t_e$ represents start and end timestamps, and $s_i$ represents the associated proposal score. We extract video features as $F_v \in \mathbb{R}^{T\times d_v}$ by C3D network \cite{tran2015c3d}, and query feature as $F_q \in \mathbb{R}^{d_q}$ by GRU network \cite{cho2014gru}.

\subsection{Proposal Module}
The core functionality of our proposal module is to construct learnable proposal representation to facilitate the follow-up processing. Proposals (segments) in a video exhibit much higher variance than object bounding boxes in an image regarding proposal scales. Therefore, it is vital to obtain fix-length proposal representation while maintaining discriminative features. To tackle this issue, we follow similar spirit as \cite{lin2019bmn} and propose a dynamic proposal sampling mechanism to efficiently process dense variable-length proposals and obtain fix-length proposal-level features by a 3D convolutional layer. As shown in Figure \ref{fig:method} (a), we first adopt a 1D convolutional layer to reduce the dimensionality while considering temporal context by setting the convolutional kernel size as 3:

\begin{equation}\label{eq:conv1d}
\begin{aligned}
\hat{F}_{v} = \textrm{Conv1d}(F_{v}) ,
\end{aligned}
\end{equation}
where $\hat{F}_v \in \mathbb{R}^{T\times \frac{d_v}{r}}$, $r$ is the reduction ratio. Then a sampling weight map is designed as \cite{lin2019bmn} followed by a 3D convolutional layer:
\begin{equation}\label{eq:dot}
\begin{aligned}
F^{\prime}_{v} = W \otimes \hat{F}_{v},
\end{aligned}
\end{equation}
\begin{equation}\label{eq:conv3d}
\begin{aligned}
F_{vp} = \textrm{Conv3d}(F^{\prime}_{v}) ,
\end{aligned}
\end{equation}
where $W \in \mathbb{R}^{T\times S\times N\times T}$ is the pre-calculated sampling weight map, $\otimes$ represents matrix multiplication. We permute dimension for obtained $F^{\prime}_v$ from $F^{\prime}_v\in \mathbb{R}^{T \times S \times N \times \frac{d_v}{r}}$ as $F^{\prime}_v\in \mathbb{R}^{N \times T \times S \times \frac{d_v}{r}}$ . After Conv3d, $F_{vp} \in \mathbb{R}^{T\times S\times d_{vp}}$ is obtained as fix-length proposal representation.
In detail, for sampling map $W$, We can rewrite it as:
\begin{equation}\label{eq:weight map}
\begin{aligned}
W \in \mathbb{R}^{(T \times S)\times (N \times T)} .
\end{aligned}
\end{equation}
The former part $(T\times S)$ represents the shape of the proposal location map, where $T$ represents possible start times of the sampled proposals, and $S$ is the number of pre-defined proposal scales. In such a way, our proposal module can cover sufficient candidate proposals in the video. The later part $(N \times T)$ can be regarded as the shape of a sampling weight map $W_{ij}\in \mathbb{R}^{N\times T}$ for each proposal. In detail, we uniformly sample $N$ temporal points for each proposal, then a $T$-dimensional sampling weight vector is generated according to their locations and further used to determine the contributions of the original $T$ features. Specifically, for a non-integer sampling point $t_n$, the weight vector $W_{i,j,n} \in \mathbb{R}^{T}$ is obtained as:

\begin{equation}\label{eq:grad_simp}
    W_{i,j,n}[t] =
    \begin{cases}
    1-dec(t_n), &\text{if } t=\lfloor t_n\rfloor\\
    dec(t_n), &\text{if } t=\lfloor t_n\rfloor+1\\
    0, &\text{otherwise}
    \end{cases} ,
\end{equation}
where $dec$ is decimal function and $\lfloor t_n \rfloor$ means the integer part of $t_n$. After the sampling mechanism described above, we feed the obtained proposal-level sampling features into the 3D convolutional layer to get proposal-level feature. The size of the 3D convolutional kernel is set to $(N \times 1 \times 1)$, which is responsible for summarizing intra-proposal information, meanwhile reducing the shape of proposal representation from $N \times d_{vp}$ to $1 \times d_{vp}$. The corresponding output can be regarded as the proposal-level feature $F_{vp} \in \mathbb{R}^{T \times S \times d_{vp}}$ and will be fed into the following attention module.

\subsection{Attention Module}
The graphical description of the attention module can be found in part (b) of Fig. \ref{fig:method}. As we expect the learnt proposal attentions can well reflect the semantic meaning of the given query, we first fuse each proposal with the given sentence by concatenating the proposal feature and the sentence feature, then obtain attention scores by cascaded convolutional layers:
\begin{equation}\label{eq:fuse}
\begin{aligned}
F_{a} = \textrm{Concat}(F_{vp}, \textrm{Tile}(F_{q})) ,
\end{aligned}
\end{equation}
\begin{equation}\label{eq:att}
\begin{aligned}
Att_p = \textrm{Softmax}
\left(\textrm{Conv2d}_2(\textrm{ReLU}(\textrm{Conv2d}_1(F_{a})))\right) .
\end{aligned}
\end{equation}
Here $F_{q}$ is the encoded query feature. Tile means duplicating the query feature along the temporal dimension ($T$) and the proposal scale dimension ($S$). Conv2d$_1$ is used to get the integrated proposal-level multi-modal feature $F_a \in \mathbb{R}^{T \times S \times d_a}$, and Conv2d$_2$ is to obtain the corresponding proposal-level attention $Att_p \in \mathbb{R}^{T\times S\times 1}$.

Note that the representations of adjacent proposals corresponding to neighboring positions in the proposal map. This makes it convenient to calculate proposal-level context. We design several structures for Conv2d$_2$ by adjusting the size of receptive field to build inter-proposal interaction. Such an interaction is expected to obtain more reliable attention scores by comparing the current proposal with its surrounding proposals. The ablation study is provided in Section ~\ref{sec:ablation}.

The learnt attention map is further used to calculate proposal-driven global feature $F_{p}^{global} \in \mathbb{R}^{1 \times d_{vp}}$ by:
\begin{equation}\label{eq:globalf}
\begin{aligned}
F_{p}^{global} = F_{vp} \otimes Att_p .
\end{aligned}
\end{equation}
We apply the attention map to summarize all the proposal-level representations. Here we use $F_{vp}$, as it corresponds to the learnt proposal features which contain no textual query information. If we apply the attentions to $F_a$ instead, the model may overfit to remember the textual query for reconstruction, and thus disrupt the meaning of the calculated attentions. We require the model to reconstruct the query by only visual information. Here $\otimes$ represents matrix multiplication. By training the reconstruction module (described in the next sub-section), this attention map is regularized to highlight the correct proposals by assigning them with higher attention scores.

\subsection{Reconstruction Module}
After obtaining proposal-driven global feature $F_{p}^{global}$, we use it to reconstruct the given query word by word with LSTM:

\begin{equation}\label{eq:lstm}
\begin{aligned}
h_m = \textrm{LSTM}\left(\textrm{Concat}(F_{p}^{global}, h_{m-1}, e_{m-1})\right) ,
\end{aligned}
\end{equation}
where $e_{m-1}$ is the GloVe word embedding vector of the $(m-1)$-th word $w_{m-1}$, and $h_m$ is the next LSTM hidden state. The probability of next word is calculated by a fully-connected layer with input $h_m$. We follow the common practice for calculating captioning loss:
\begin{equation}\label{eq:loss_p}
\begin{aligned}
L_{p}^{cap} = -\frac{1}{M}\sum\limits_{m=1}^M logP\left(w_m|F_{p}^{global}, h_{m-1}, w_1, ..., w_{m-1}\right) .
\end{aligned}
\end{equation}
The captioning loss is optimized to reconstruct the given query by the attended global feature. The proposal-level attentions will be trained to highlight the proposals that can better express the given sentence.

\subsection{Multi-Level Extension}
To compensate fine-grained video information, we extend the framework to a multi-level structure by adding a clip-level branch which reconstructs the input query by attended global video representation driven by clip-level features. The clip-level features are similarly obtained by first performing feature dimensional reduction with 1D convolution followed by the sampling procedure. Here the number of scales is 1, and thus the sampling weight matrix $W\in \mathbb{R}^{T\times 1 \times N\times T}$. The clip-level features after performing 3D convolution are then $F_{vc}\in \mathbb{R}^{T\times 1 \times d_{vc}}$. The clip-level attentions are similarly calculated as:
\begin{equation}\label{eq:fuse}
\begin{aligned}
Att_{c} = \textrm{Softmax}\left(\textrm{Conv2d}(\textrm{Concat}(F_{vc}, \textrm{Tile}(F_{q})))\right) ,
\end{aligned}
\end{equation}
\begin{equation}\label{eq:globalf}
\begin{aligned}
F^{global}_c = F_{vc} \otimes Att_c .
\end{aligned}
\end{equation}

The reconstruction module for the clip-level branch is the same as the proposal-level branch, and we share the same LSTM paramters. We apply multi-task loss with a hyper-parameter $\lambda$ to jointly train the two branches.
\begin{equation}\label{eq:loss_c}
\begin{aligned}
L^{cap}_c = -\frac{1}{M}\sum\limits_{m=1}^M logP\left(w_m|F_{c}^{global}, h^{\prime}_{m-1}, w_1, ..., w_{m-1}\right) ,
\end{aligned}
\end{equation}
\begin{equation}\label{eq:loss}
\begin{aligned}
L^{cap} = L^{cap}_p + \lambda L^{cap}_c .
\end{aligned}
\end{equation}

During inference, we utilize clip-level attentions to refine the confidence scores of the proposals. We expect the clip-level attentions should highlight the internal point of the target proposals (actionness). Moreover, we assume generally the middle actionness of a proposal should obtain relatively higher attention coefficient. Therefore, we modify the final score for a proposal by integrating the original proposal attention with the clip-level attention if the clip is located at the central point of the proposal:
\begin{equation}\label{eq:infer_ensemble}
\begin{aligned}
Att^{\prime}_p(i) = Att_p(i) + \epsilon Att_c(j) , \text{if } j = T_i + \frac{1}{2}S_i ,
\end{aligned}
\end{equation}
 where $\epsilon$ is a hyper-parameter to balance the contributions from the proposal-level attentions and the clip-level attentions. $T_i$ is the start time of $i$-th proposal, $S_i$ is the corresponding temporal scale.

\section{Experiments}
In this section, we evaluate our method on two benchmark datasets: Charades-STA dataset \cite{gao2017tall} and ActivityNet-Captions dataset. \cite{krishna2017dense}. In the following, we will describe the detailed experimental settings for each dataset, offer comparisons with existing state-of-the-art approaches, and discuss the effect of our proposed methods by ablation study.

\subsection{Datasets}
Charades-STA \cite{gao2017tall} and ActivityNet-Captions \cite{krishna2017dense} are two commonly-adopted datasets for evaluating temporal video grounding performance.

\vspace{3pt}
\noindent\textbf{Charades-STA} contains around 10k videos with temporal activity annotations (from 157 indoor activity categories) and multiple video-level descriptions. Since segment-level sentence annotations are not provided in the original Charades dataset, \cite{gao2017tall} designed a semi-automatic way to generate temporal annotation for each sentence using off-the-shelf NLP algorithms and the annotations were further verified manually. The released dataset is composed of 12,408 sentence-clip pairs for training and 3,720 pairs for testing.

\vspace{3pt}
\noindent\textbf{ActivityNet-Captions} is a large-scale dataset including 20k videos, and contains around 100k densely annotated proposal-level temporal locations associated with manually-written sentences. The videos generally cover a wide rage of real-life activities, including climbing mountain, kayak, etc. Since the test split is withheld for competition, we merge the two validation subsets ``val\_1'', ``val\_2'' as our test split, as the two validation sets correspond to the same videos \cite{krishna2017dense}. The numbers of query-segment pairs for train/test split are thus 37421 and 34536.

\subsection{Experimental Settings}
In this section we describe the detailed experimental settings including feature extraction, detailed hyper-parameter settings, compared approaches, and evaluation metric.

\vspace{3pt}
\noindent\textbf{Implementation Details.}
For fair comparison, C3D \cite{tran2015c3d} features are extracted to calculate $F_{vp}$. For each second we uniformly sample 16 frames as input to C3D, and obtain a 4096-dimentional visual feature from fc6 layer. Each word from the query is represented by GloVe \cite{pennington2014glove} word embedding vector pre-trained on Common Crawl. For sentences we obtain global textual features by feeding the extracted GloVe word embedding vectors into GRU. For Charades-STA, the video temporal length $T$ is set as 32, and the dimensional reduction ratio $r$ is set as 8. We use $W \in \mathbb{R}^{32 \times 6 \times 4 \times 32}$ to generate 6-scale proposals in the range of $[6, 12]$. For ActivityNet-Captions, the video temporal length $T$ is set as 128. Each temporal unit corresponds to 2-second snippet. The reduce scale $r$ is 32. The sample weight map is set as $W \in \mathbb{R}^{128 \times 64 \times 4 \times 128}$. Considering the large overlap among long proposals, we adopt sparse sampling by proposal masking. Specifically, for proposals with the same scale $s$, instead of densely sampling proposals with stride of 1, we set a dynamic stride as $\frac{1}{4}s$. Also, any proposal out of video range is masked out. The hyper-parameter $\epsilon$ used in multi-level inference is set as 0.1 for Charades-STA and 0.3 for ActivityNet-Captions.

\vspace{3pt}
\noindent\textbf{Compared Methods.}
We evaluate our proposed model by comparing against the following state-of-the-art methods, including both fully-supervised methods and weakly-supervised methods.
The compared fully-supervised methods are as follows. \textbf{Random}: The candidate proposals are randomly ranked. \textbf{VSA-RNN} \cite{karpathy2015deep} and \textbf{VSA-STV} \cite{karpathy2015deep}: visual-semantic alignment with LSTM or skip-thought vectors \cite{kiros2015skip}. \textbf{CTRL} \cite{gao2017tall}:  a cross-model temporal regression localizer. \textbf{ABLR} \cite{yuan2019ablr}: attention based location regression approach to directly predict segment locations. \textbf{QSPN} \cite{xu2019multilevel}: a two-stage method (generation + reranking) exploiting re-captioning. \textbf{TGN} \cite{chen2018tgn}: Temporal GroundNet. \textbf{SCDM} \cite{yuan2019scd}: a model with a SSD-based semantic conditioned dynamic modulation mechanism.
The compared weakly-supervised methods are as follows. \textbf{TGA} \cite{mithun2019tga}: utilizes clip-level alignment by text guided attention. \textbf{WSLLN} \cite{gao2019wslln}: a weakly-supervised language localization network. \textbf{wMAN} \cite{tan2019wman}: a weakly-supervised moment alignment network. \textbf{SCN} \cite{lin2019scn}: semantic completion network. \textbf{CTF} \cite{chen2020look}: a two-stage method in coarse-To-fine manner.
For fair comparison, we follow the same settings regarding visual and textual features. We compare with the best reported results in their papers.

\vspace{3pt}
\noindent\textbf{Evaluation Metrics.}
Following prior work, we mainly adopt ``R@$\mathit{N}$, IoU=$\mathit{\theta}$'' and ``mIoU'' as the evaluation metrics. ``R@$\mathit{N}$, IoU=$\mathit{\theta}$'' represents the percentage of top $\mathit{N}$ results that have at least one segment with higher IoU (Intersection over Union) than $\mathit{\theta}$. ``mIoU'' computes the average IoU of top-1 result with ground-truth segment over all testing queries. Since higher IoU represents more precise localization, recall on higher IoUs should be more meaningful than recall on lower IoUs.

\subsection{Comparison with State-of-the-Arts}
In this section, we compare our method with both fully-supervised methods and the existing weakly-supervised methods. The proposed MARN outperforms the existing weakly-supervised video grounding approaches with a clear margin, and meanwhile shows competitive results than some fully-supervised methods. The achieved promising results indicate that it is plausible and meaningful to carry on deeper investigation on the weakly-supervised video grounding.

\subsubsection{Charades-STA}
Following prior work, we report results for IoU $\in \{0.7,0.5,0.3\}$ and Recall @$ \{1, 5\}$ for Charades-STA dataset, as in Table \ref{table:charades}. It is obvious that the grounding performance degenerates as IoU gets higher, for both fully-supervised approaches and weakly-supervised approaches. Compared to CTRL, QSPN, and SCDM, VSA-RNN and VSA-STV can only achieve very limited performance as they do not exploit any contextual information for localization. Our model achieves a recall of 14.81\% for the metric ``R@1,IoU=0.7'', surpassing VSA-RNN, VSA-STV, and CTRL. Compared to other weakly-supervised methods, we also achieve better results with a clear margin. Specifically, our method outperforms all the existing weakly-supervised video grounding methods for ``R@1''. Noticeably, compared to the previous best performing approach (wMAN), our method obtains more performance gain on the important metric ``R@1,IoU=0.7'' than other metrics, indicating that our method can give more precise grounding results. For ``R@5'', we only obtain competitive performance compared to SCN. However, on ``R@1'', the performance of SCN is much inferior to ours (e.g., 9.97 \emph{vs.} 14.81 on ``R@1,IoU=0.7''). We also notice that although wMAN achieves much higher ``R@1, IoU=0.7'' than SCN (13.71 \emph{vs.} 9.97), it also obtains slightly lower ``R@5,IoU=0.7''.
\vspace{-10pt}
\begin{table}
       \scriptsize
		\centering
        \caption{Comparison of different methods on Charades-STA.}
        \vspace{-5pt}
        \scalebox{0.98}{

		\begin{tabular}{lcccccccc}
            \toprule
            \multirow{2}{*}{\textbf{Method}} & {\textbf{Training}}
            &\multicolumn{3}{c}{\textbf{R@1}} & \multicolumn{3}{c}{\textbf{R@5}}\\
            \cmidrule(r){3-5} \cmidrule(r){6-8}
            &{\textbf{Supervision}}& IoU=0.7 & IoU=0.5 & IoU=0.3 & IoU=0.7 & IoU=0.5 & IoU=0.3\\
            \midrule
            Random & {--} & 3.39 & 10.50 & 20.12 & 14.98 & 37.57 &68.42\\
            VSA-RNN \cite{gao2017tall}& {\textbf{Fully}} & 4.32 & 10.50 & -- & 20.21 & 48.43 & --\\
            VSA-STV \cite{gao2017tall} & {\textbf{Fully}} & 5.81 & 16.91 & -- & 23.58 & 53.92 & --\\
            CTRL \cite{gao2017tall} & {\textbf{Fully}} & 7.15 & 21.42 & -- & 26.91 & 59.11 & --\\

            QSPN \cite{xu2019multilevel} & {\textbf{Fully}} & 15.80 & 35.60 & 54.70 & 45.40 &79.40  &95.60\\
            SCDM \cite{yuan2019scd}& {\textbf{Fully}} & 33.43& 54.44 & --& 58.08 & 74.43 & -- \\
            \midrule
            \midrule
            TGA \cite{mithun2019tga} & {\textbf{Weakly}} & 8.84 & 19.94 & 32.14 & 33.51 & 65.52 & 86.58\\
            SCN \cite{lin2019scn} & {\textbf{Weakly}} & 9.97 & 23.58 & 42.96 & \textbf{38.87} & 71.80 & \textbf{95.56}\\
            CTF \cite{chen2020look} & {\textbf{Weakly}} & 12.90 & 27.30 & 39.80 & -- & -- & --\\
            wMAN \cite{lin2019scn} & {\textbf{Weakly}} & 13.71 & 31.74 & 48.04 & 37.58 & \textbf{72.17} & 89.01\\
            \midrule
            \textbf{Ours} & {\textbf{Weakly}}
            &\textbf{14.81} & \textbf{31.94} & \textbf{48.55} & 37.40 & 70.00 & 90.70\\
            \bottomrule
        \end{tabular}}
        \label{table:charades}
	\end{table}
	
\vspace{-30pt}
	
\begin{table}
       \scriptsize
		\centering
        \caption{Comparison of different methods on ActivityNet-Captions.}
        \vspace{-5pt}
        \scalebox{0.98}{
		\begin{tabular}{lcccccc}
            \toprule
            \multirow{2}{*}{\textbf{Method}} & {\textbf{Training}}
            &\multicolumn{2}{c}{\textbf{R@1}} & \multicolumn{2}{c}{\textbf{R@5}}\\
            \cmidrule(r){3-4} \cmidrule(r){5-6}
            &{\textbf{Supervision}}& IoU=0.5 & IoU=0.3 & IoU=0.5 & IoU=0.3 \\
            \midrule
            Random & --& 7.73 & 18.64 & 29.49 & 52.78\\
            VSA-RNN \cite{gao2017tall}& {\textbf{Fully}}
            & 23.43 & 39.28& 55.52 & 70.84\\
            VSA-STV \cite{gao2017tall} & {\textbf{Fully}}
            &24.01 & 41.71& 56.62 & 71.05\\
            CTRL \cite{gao2017tall} & {\textbf{Fully}}
            & 14.00 & 28.70 & -- & --\\
            ABLR \cite{yuan2019ablr} & {\textbf{Fully}}
            & 36.79 &55.67  &--&--\\
            TGN \cite{chen2018tgn}& {\textbf{Fully}}
            &27.93 &43.81 & 44.20 & 54.56 \\
            QSPN \cite{xu2019multilevel} & {\textbf{Fully}}
            & 27.70 & 45.30 & 59.20 & 75.70\\
            SCDM \cite{yuan2019scd}& {\textbf{Fully}}
            & 36.75& 54.80& 64.99 &77,29\\
            \midrule
            \midrule
            WSLLN \cite{gao2019wslln} & {\textbf{Weakly}}
            & 22.70 & 42.80 & -- & -- \\
            SCN \cite{lin2019scn} & {\textbf{Weakly}}
            & 29.22 & \textbf{47.23} & 55.69 & 71.56\\
            CTF \cite{chen2020look} & {\textbf{Weakly}}
            & 23.60 & 44.30 & -- & --\\
            \midrule
            \textbf{Ours} & {\textbf{Weakly}}
            & \textbf{29.95} & 47.01 & \textbf{57.49} & \textbf{72.02}\\
            \bottomrule
        \end{tabular}}
		\label{table:activity}
		\vspace{-10pt}
	\end{table}
	
\subsubsection{ActivityNet-Captions}
Following the existing approaches, we evaluate our method using IoU $\in \{0.5,0.3\}$ and Recall @$ \{1, 5\}$. The results are shown in Table~\ref{table:activity}. Consistent with the findings from Charades-STA, VSA-RNN and VSA-STV perform inferiorly to other approaches on ActivityNet-Captions dataset. Noticeably, our model achieves even better performance than some fully-supervised approaches, including CTRL, TGN, QSPN. Among the previous weakly-supervised approaches (WSLL, SCN, CTF), SCN performs the best. Compared with SCN, our model obtains superior performance on almost all the metrics. For higher IoUs (IoU=0.5), we obtain more performance gain. This again gives evidence that the proposed method consistently produces more precise localization. By taking both datasets into consideration, our model substantially outperforms SCN on Charades-STA and clearly outperforms SCN on ActivityNet-Captions, which provides strong evidence on the superiority of our method.
	
\subsection{Ablation Study}
\label{sec:ablation}
The reconstruction module provides loss to optimize the whole network, which is non-trivial to be ablated. The proposal module and the attention module are two fundamental components of our framework. We would like to compare the proposed modules with existing techniques for weakly-supervised grounding. The ablation study of the two modules is provided in Table \ref{table:ablation_charades} and Table \ref{table:ablation_activity}. In both tables we first conduct ablation study on the attention module (No. 1, 2, 3), followed by verifying different 1D convolution (No. 2, 4) of the proposal module, and finally conduct ablation study on temporal representation (TempR) of the proposal module. Each experiment is performed based on the previous best setting. Further, we would like to compare the contribution from the clip-level branch for both training and inference. The corresponding multi-level experiments are also conducted and the results are provided in Table \ref{table:ablation_multilevel}.

\begin{table}
       \scriptsize
		\centering
		\renewcommand\arraystretch{1.1}
        \caption{Ablation study on Charades-STA for  the proposal module and the attention module.}
        \scalebox{0.88}{
		\begin{tabular}{cccccccccccc}
            \toprule
            \multirow{2}{*}{\textbf{No.}} &\multicolumn{3}{c}{\textbf{Proposal Module}}
            &\multicolumn{2}{c}{\textbf{Attention Module}}
            &\multicolumn{3}{c}{\textbf{R@1}}  &\multicolumn{3}{c}{\textbf{R@5}}\\
            \cmidrule(r){2-4} \cmidrule(r){5-6} \cmidrule(r){7-9} \cmidrule(r){10-12}
            &{Conv1d}&{TempN}&{TempR}&\multicolumn{2}{c}{Conv2d}& IoU=0.7 & IoU=0.5 &IoU=0.3& IoU=0.7& IoU=0.5 & IoU=0.3\\
            \midrule

            (1)& k=1 & All & AvgPool &  \multicolumn{2}{c}{k=(1$\times$1)}
            &7.94 &20.10 & 34.19 &27.40 & 47.17 & 58.35\\
            (2)& k=1 & All & AvgPool &  \multicolumn{2}{c}{k=(3$\times$3)}
            &8.41 &23.20 & 38.28 &27.88 & 48.08 & 64.33\\
            (3)& k=1 & All & AvgPool &  \multicolumn{2}{c}{k=(3$\times$3),$\times$2}
            &7.19 &21.28 &36.15 &25.27 &46.36 &63.15\\

            (4)& k=3 & All & AvgPool & \multicolumn{2}{c}{k=(3$\times$3)}
            & 12.63 &26.89 &39.95 &31.57 &57.17 &\textbf{75.86}\\

            (5)& k=3 & All & MaxPool &  \multicolumn{2}{c}{k=(3$\times$3)}
            &9.48 &20.23 &30.06 &32.46 &\textbf{57.87} &73.98 \\
            (6)& k=3 & All & LSTM &  \multicolumn{2}{c}{k=(3$\times$3)}
            & 13.23 & 27.86 &40.87 &31.12 &53.90 &71.39\\
            (7)& k=3 & 4 & Conv3d & \multicolumn{2}{c}{k=(3$\times$3)}
            &\textbf{14.39} & \textbf{32.00} &\textbf{43.53} & \textbf{33.73} & 55.44 & 67.91\\
            \bottomrule
        \end{tabular}}
		\label{table:ablation_charades}
		
	\end{table}

\begin{table}
       \scriptsize
		\centering
		\renewcommand\arraystretch{1.1}
        \caption{Ablation study on ActivityNet-Captions for the proposal module and the attention module.}
        \scalebox{0.99}{
		\begin{tabular}{cccccccccc}
            \toprule
            \multirow{2}{*}{\textbf{No.}} &\multicolumn{3}{c}{\textbf{Proposal Module}}
            &\multicolumn{2}{c}{\textbf{Attention Module}}
            &\multicolumn{2}{c}{\textbf{R@1}}  &\multicolumn{2}{c}{\textbf{R@5}}\\
            \cmidrule(r){2-4} \cmidrule(r){5-6} \cmidrule(r){7-8} \cmidrule(r){9-10}
            &{Conv1d}&{TempN}&{TempR}&\multicolumn{2}{c}{Conv2d}& IoU=0.5 &IoU=0.3& IoU=0.5 & IoU=0.3\\
            \midrule

            (1)& k=1 & All & AvgPool &  \multicolumn{2}{c}{k=(1$\times$1)}
            & 17.82 & 32.00 &35.14 & 48.94\\
            (2)& k=1 & All & AvgPool &  \multicolumn{2}{c}{k=(3$\times$3)}
            & 18.16 & 35. 14 & 43.40 & 62.74\\
            (3)& k=1 & All & AvgPool &  \multicolumn{2}{c}{k=(3$\times$3),$\times$2}
            & 17.45 & 31.72 & 35.50 & 50.24\\

            (4)& k=3 & All & AvgPool & \multicolumn{2}{c}{k=(3$\times$3)}
            & 20.40 & 36.74 & 39.85 & 48.06\\

            (5)& k=3 & All & MaxPool &  \multicolumn{2}{c}{k=(3$\times$3)}
            & 18.40 &33.49& \textbf{47.52} & \textbf{65.33}\\
            (6)& k=3 & All & LSTM &  \multicolumn{2}{c}{k=(3$\times$3)}
            & 25. 06 & 38.95& 39.70 & 57.20 \\
            (7)& k=3 & 4 & Conv3d & \multicolumn{2}{c}{k=(3$\times$3)}
            & \textbf{29.48} & \textbf{47.17} & 46.34 & 64.39 \\
            \bottomrule

            \hline
        \end{tabular}}
		\label{table:ablation_activity}
	\end{table}

\subsubsection{Effect of Attention Module}
The receptive field for each element in the proposal map can be expanded by using larger 2D convolutional kernels or stacking multiple kernels. As shown in the experiments with No. (1)(2)(3) in Table \ref{table:ablation_charades} and Table \ref{table:ablation_activity}, a proper receptive field does affect the final performance. Among them, the experiment for 2D convolution with kernel size of $3 \times 3$ achieves the best performance. It indicates that visual context modeling with proper receptive field is important for grounding textual queries.

\subsubsection{Effect of Proposal Module}
Based on the best practice from the attention module (No. (2)), we compare different kernel sizes for the 1D convolutional layer, which is responsible for performing feature dimension reduction. Comparing No. (4) with No. (2), we find that expanding temporal kernel size from 1 to 3 helps obtain better grounding results. Based on the setting of No. (4), we further verify different temporal representation approaches. We compare the 3D convolution with the common-adopted average pooling (AvgPool), max pooling (MaxPool), and LSTM. The corresponding results are shown in No. (4), (5), (6), and (7). In both Charades-STA and ActivityNet-Captions dataset, 3D convolution operation consistently outperforms the existing pooling techniques including average pooling, max pooling and LSTM. In Charades-STA, we obtain 1.16\% perform gain and we obtain over 4\% gain for the important metric ``R@1,IoU=0.7'', compared to the LSTM baseline. Also, our model is much more efficient than LSTM as LSTM cannot parallelly process the features.

\subsubsection{Effect of Multi-Level Extension}
We next verify the effectiveness of the proposed clip-level branch. As shown in Table \ref{table:ablation_multilevel}, when only applied at the training phase, the clip-level branch has already improved the localization precision from 14.39 to 14.70 for the ``R@1,IoU=0.7'' metric. The reason could be that the parameters of 3D convolution and the reconstruction for the two branches are shared, and thus the clip-level branch can help regularize the network optimization during the training stage. When ensembled at the inference stage using Eq. \ref{eq:infer_ensemble}, the final results can be further improved.

\begin{table}
       \scriptsize
		\centering
        \caption{Ablation study on Charades-STA and ActivityNet-Caption for Multi-level progress.}
        \scalebox{0.9}{
		\begin{tabular}{cccccccccccc}
            \toprule
            \multicolumn{2}{c}{\textbf{Multi-Level}}
            &\multicolumn{6}{c}{\textbf{Charades-STA}}
            &\multicolumn{4}{c}{\textbf{ActivityNet-Captions}}\\
            \cmidrule(r){3-8}\cmidrule(r){9-12}
            \multirow{2}{*}{\textbf{Train}}
            &\multirow{2}{*}{\textbf{Infer}}
            & R@1 & R@1& R@1 & R@5 & R@5 & R@5
            & R@1 & R@1& R@5 & R@5 \\
            & & IoU=0.7 & IoU=0.5 & IoU=0.3
            & IoU=0.7 & IoU=0.5 & IoU=0.3
            & IoU=0.5 & IoU=0.3
            & IoU=0.5 & IoU=0.3 \\
            \midrule
             &  &14.39 & 32.00 &43.53 &33.73 & 55.44 & 67.91
            &29.48 & \textbf{47.17} &46.34& 64.39 \\
            \checkmark & &14.70 & \textbf{32.07} & 47.33& 36.79 & 69.30 & \textbf{91.00}
            &29.72 & 46.34 & 57.43 & \textbf{72.76}\\

            \checkmark &\checkmark& \textbf{14.81} & 31.94 & \textbf{48.55} & \textbf{37.40} & \textbf{70.00} &90.70
            &\textbf{29.95} & 47.01 & \textbf{57.49} & 72.02 \\
            \bottomrule
        \end{tabular}}
		\label{table:ablation_multilevel}
	\end{table}

\subsection{Qualitative Analysis}
\begin{figure}
\centering
\includegraphics[width=0.85\textwidth]{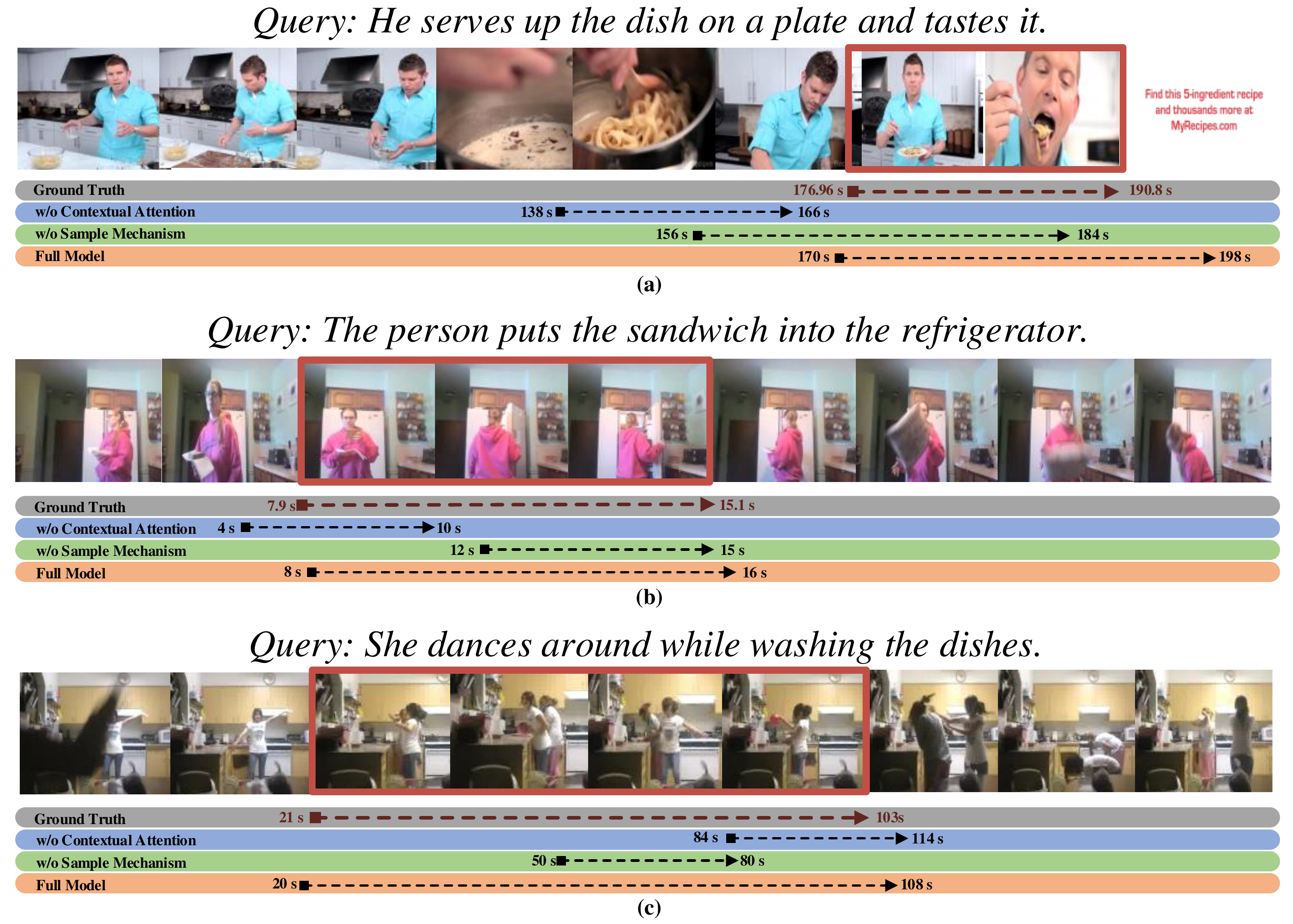}
\caption{Qualitative prediction examples. The first and the third examples are from ActivityNet-Captions dataset. The second example is from Charades-STA dataset. The top-1 result from each model is shown.}
\label{fig:example}
\end{figure}

We provide qualitative analysis to validate the effectiveness of the proposed MARN. We compare our full model with two variants: ``w/o Contextual Attention'', and ``w/o Sample Mechanisum''. The first variant replaces the $3 \times 3$ 2D convolutional kernels with $1 \times 1$ kernel. The second variant removes the sampling mechanism and uses the plain average pooling instead. As shown in Figure \ref{fig:example}, the full model shows better performance as it leverages sampling mechanism and 3D convolution to learn better proposal representation (intra-proposal interaction) and exploits language-driven proposal context (inter-proposal interaction) to learn a task-oriented attention map.

\section{Conclusion}

We proposed a novel multi-level attentional reconstruction model (MARN) in this paper, which leverages both intra-proposal and inter-proposal interaction to learn a language-driven attention map, which can directly score and rank the candidate proposals at the inference stage. By constructing query reconstruction feedback, the whole model can be trained in an end-to-end manner. The promising experimental results obtained on two widely-used datasets demonstrated the superiority of our model.

\section{Supplementary Material}
\subsection{Visualization for Learnt Attention Maps}
In our main paper, we leverage the idea of attentional reconstruction for directly scoring candidate segments (proposals) by learnt proposal-level attention. In this material we visualize the learnt attention maps and compare our full model with its two variants to qualitatively verify the superiority of our proposed MARN.

\begin{figure}
\centering
\includegraphics[width=0.95\textwidth]{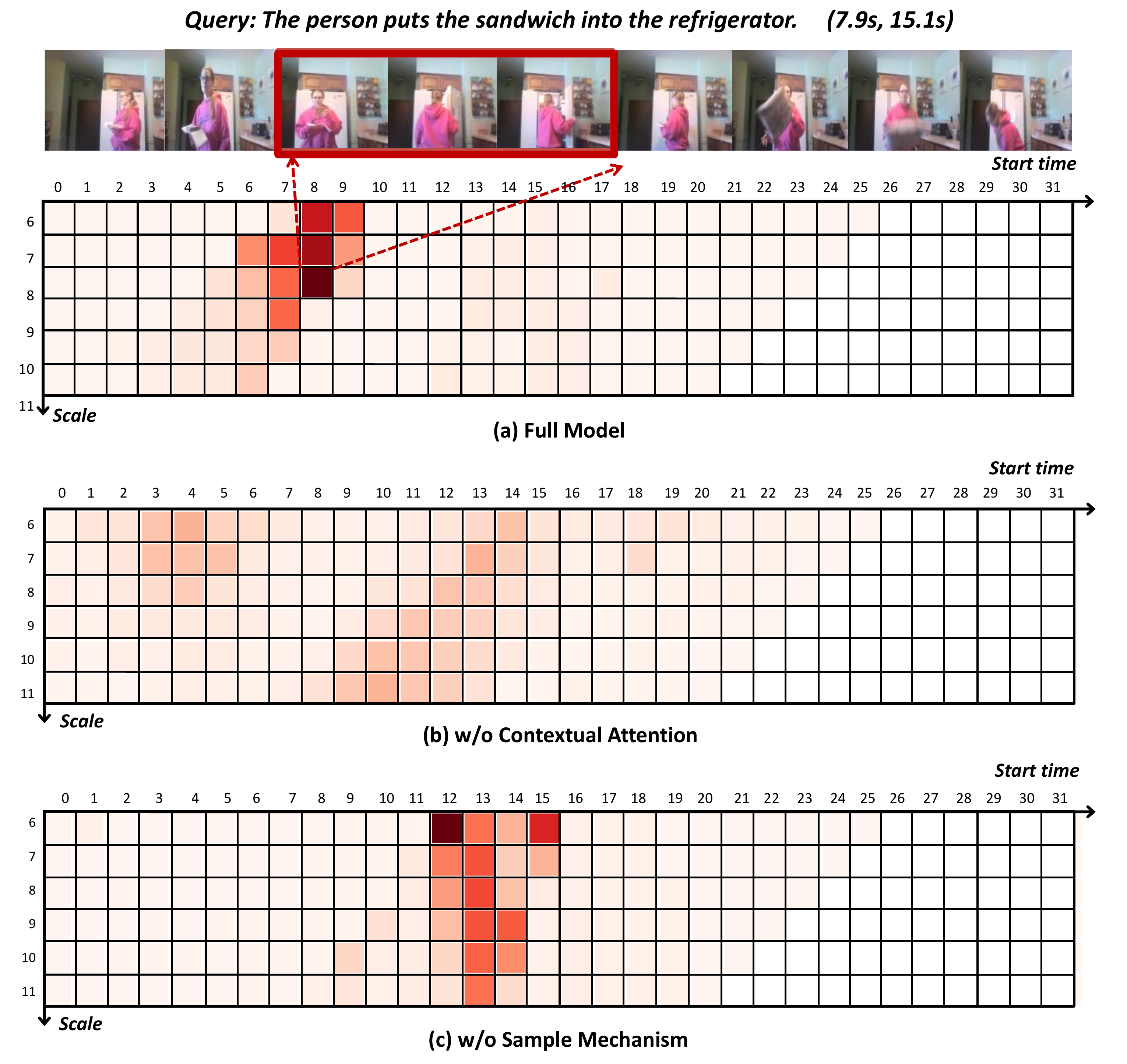}
\caption{A prediction case on Charades-STA dataset. According to the attention map, the proposal predicted by our full model is located at (8s, 16s).}
\label{fig:example_ca1}
\end{figure}

\begin{figure}
\centering
\includegraphics[width=0.95\textwidth]{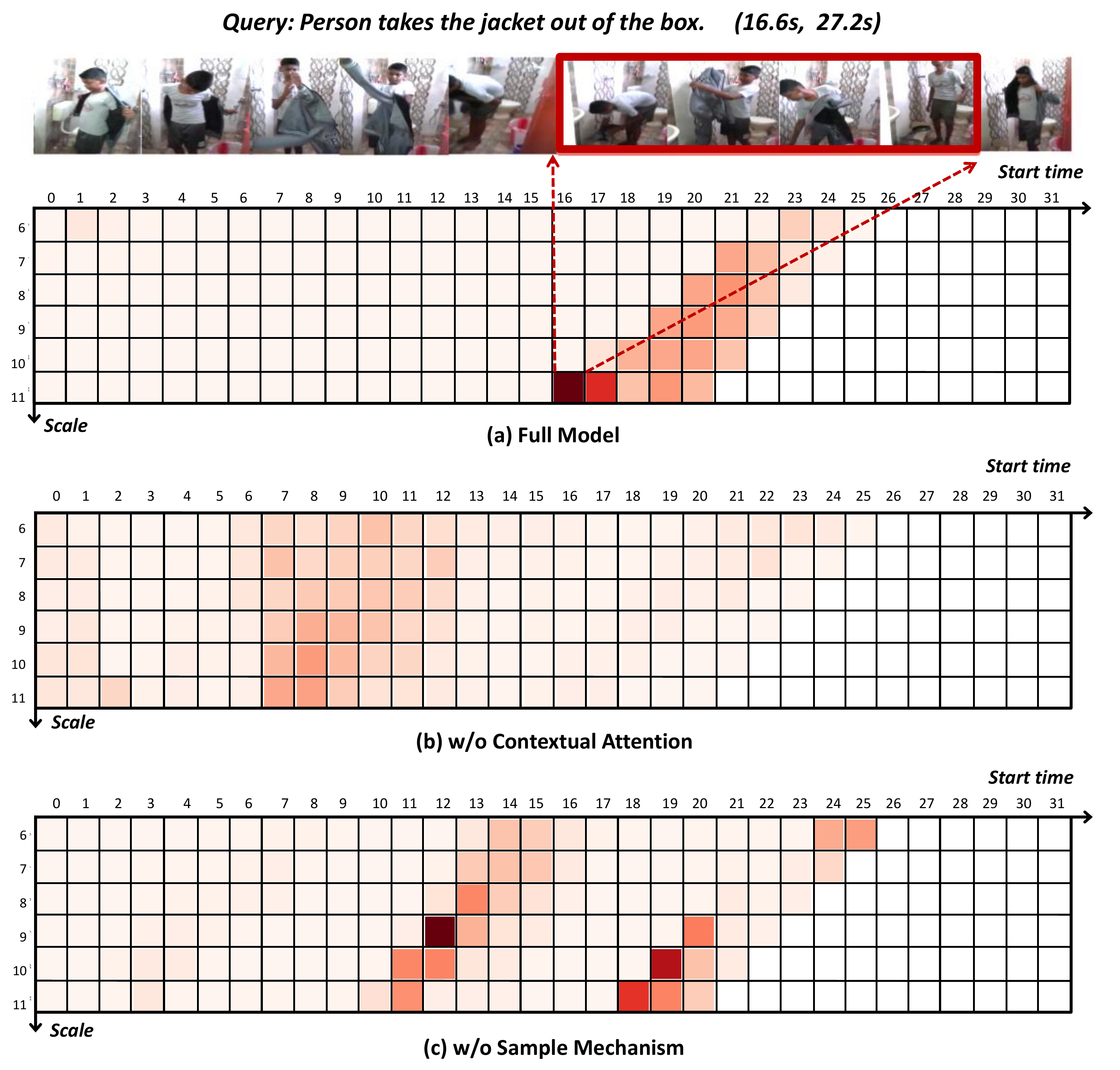}
\caption{A prediction case on Charades-STA dataset. According to the attention map, the proposal predicted by our full model is located at (16s, 28s).}
\label{fig:example_ca2}
\end{figure}

\begin{figure}
\centering
\includegraphics[width=0.88\textwidth]{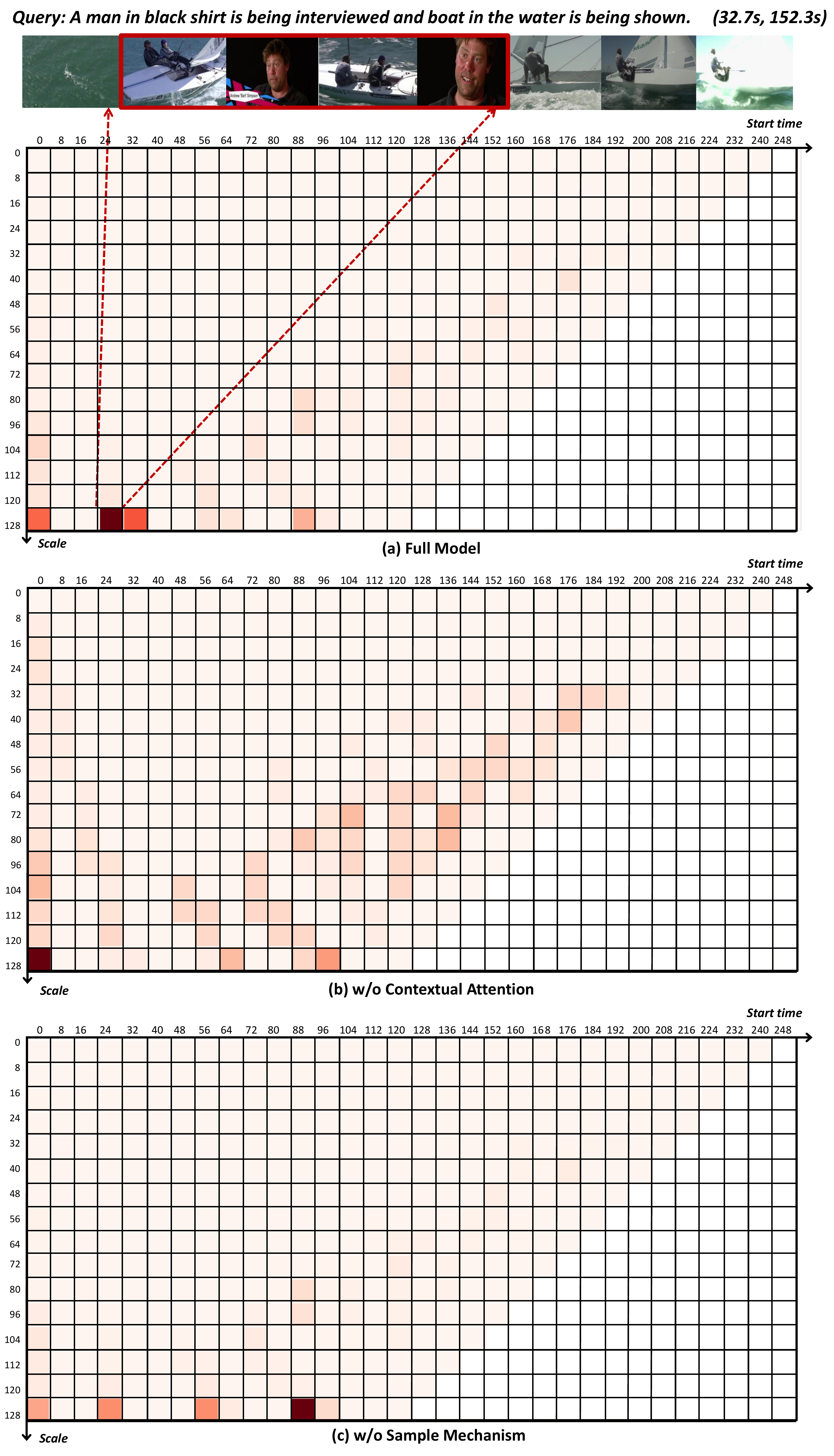}
\caption{A prediction case on ActivityNet-Captions dataset. According to the attention map, the proposal predicted by our full model is located at (30s, 152s).}
\label{fig:example_anc1}
\end{figure}

\begin{figure}
\centering
\includegraphics[width=0.88\textwidth]{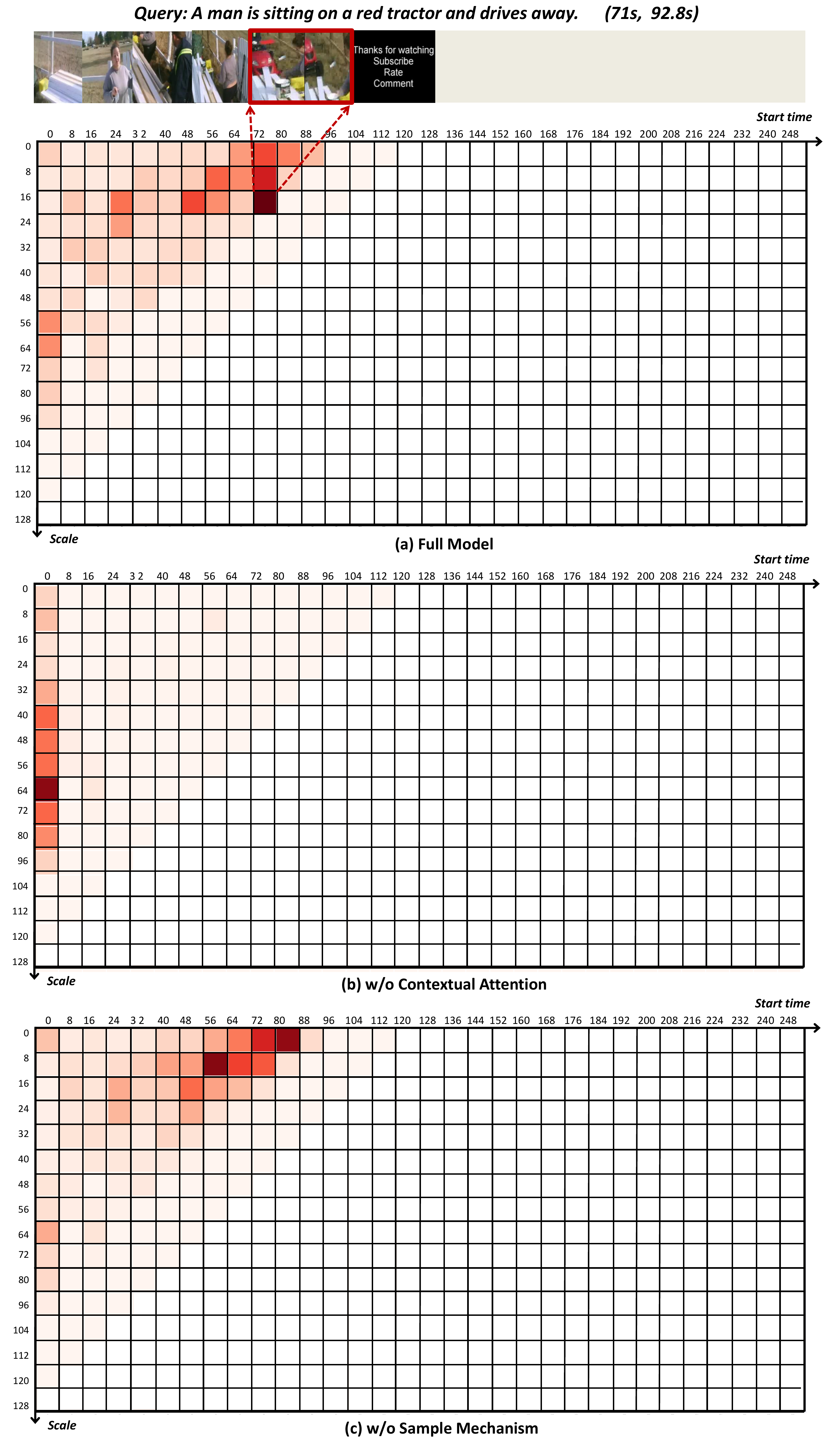}
\caption{A prediction case on ActivityNet-Captions dataset. According to the attention map, the proposal predicted by our full model is located at (72s, 92s).}
\label{fig:example_anc2}
\end{figure}

We present four prediction results together with associated visualized attention maps for Charades-STA (Fig. \ref{fig:example_ca1} and \ref{fig:example_ca2}) and ActivityNet-Captions datasets (Fig. \ref{fig:example_anc1} and \ref{fig:example_anc2}). For better visualization of videos on ActivityNet-Captions dataset with long duration, we average the attention scores from neighbouring $4 \times 4$ proposals and correspondingly adjust the coordinates of the attention maps. As shown in Fig. \ref{fig:example_ca1}, \ref{fig:example_ca2}, \ref{fig:example_anc1} and \ref{fig:example_anc2}, the full model of MARN achieves the best grounding performance, indicating that both the sampling mechanism and contextual attention modeling contribute greatly to our final model.

\clearpage
%
%

\end{document}